\documentclass[11pt,a4paper]{article}
\usepackage[hyperref]{eacl2021}
\usepackage{times}
\usepackage{latexsym}

\usepackage{microtype}

\aclfinalcopy 
\def\aclpaperid{269} 

\usepackage[linesnumbered, ruled]{algorithm2e}
\usepackage[T1]{fontenc}
\usepackage{dsfont}
\usepackage{colortbl}
\usepackage{xcolor}
\usepackage{enumitem}

\usepackage{mdframed}
\usepackage{booktabs}
\usepackage{graphicx}

\usepackage[normalem]{ulem}

\usepackage{xcolor}
\usepackage{soul}
\colorlet{soulred}{red!30}
\sethlcolor{soulred}%
\usepackage{amsmath,amsthm,amsfonts,amssymb,bm}

\usepackage{framed}
\usepackage{varioref}
\usepackage{float}
\usepackage{multirow}
\usepackage{dashrule}
\usepackage{url}
\usepackage{pifont}
\newcommand{\dygie}{\textsc{DyGIE++}}
\newcommand{\perpind}{\textsc{PerpInd}}
\newcommand{\perporg}{\textsc{PerpOrg}}
\newcommand{\target}{\textsc{Target}}
\newcommand{\victim}{\textsc{Victim}}
\newcommand{\weapon}{\textsc{Weapon}}
\newcommand{\bombing}{bombing}

\title{GRIT: Generative Role-filler Transformers \\ 
for Document-level Event Entity Extraction
}

\author{Xinya Du \ \ \quad \quad Alexander M. Rush \ \ \quad \quad Claire Cardie\\
  Department of Computer Science\\
  Cornell University \\
  {\tt \{xdu, cardie\}@cs.cornell.edu} \\
  {\tt arush@cornell.edu}
  }
  
\date{}

\begin{document}
\maketitle

\begin{abstract}
We revisit the classic problem of document-level role-filler entity extraction (REE) for template filling.
We argue that sentence-level approaches are ill-suited to the task
and introduce a generative transformer-based encoder-decoder framework (GRIT) that is
designed to model context at the document level: 
it can make extraction decisions across sentence boundaries; 
is \textit{implicitly} aware of noun phrase coreference structure, 
and has the capacity to respect cross-role dependencies in the template structure.
We evaluate our approach on the MUC-4 dataset, and show that our model performs substantially better than prior work. We also show that our modeling choices contribute to model performance, e.g., by implicitly capturing linguistic knowledge such as recognizing coreferent entity mentions. 
\end{abstract}

\section{Introduction}

\begin{figure}[!ht]
\centering
\resizebox{\columnwidth}{!}{
\includegraphics{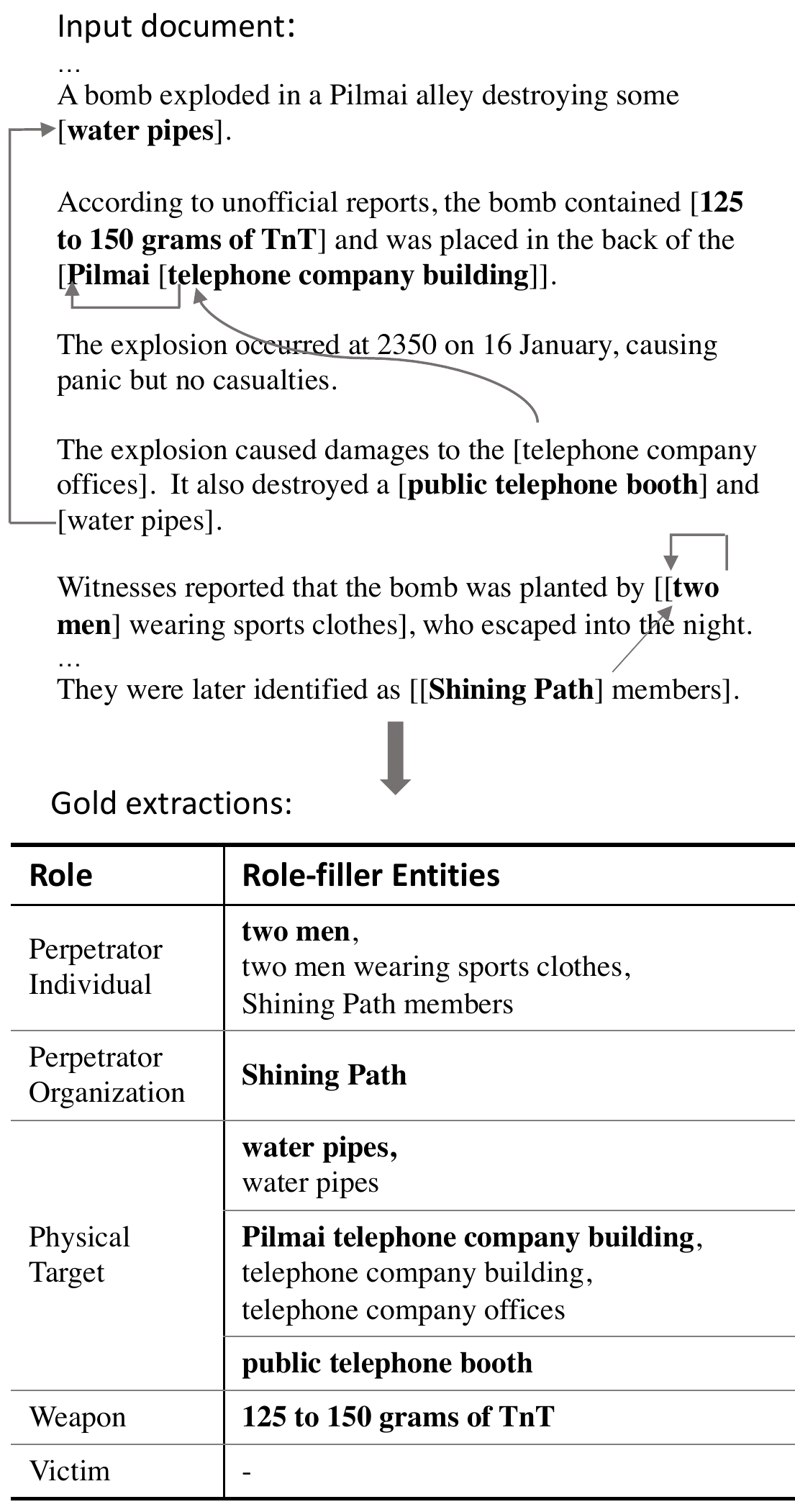}
}
\caption{Role-filler entity extraction (REE). The first mention of each role-filler entity is bold in the table and document. The arrows denote coreferent mentions.}
\label{fig:task}
\end{figure}

Document-level template filling~\cite{sundheim-1991-overview, sundheim-1993-message, grishman-sundheim-1996-design} is a classic problem in information extraction (IE) and NLP~\cite{jurafskyspeech}. 
It is of great importance for automating many real-world tasks, such as event extraction from newswire~\cite{sundheim-1991-overview}. The complete task is generally tackled in two steps. The first step detects events in the article and assigns templates to each of them (template recognition); the second step performs role-filler entity extraction (REE) for filling in the templates.
In this work we focus on the \textbf{role-filler entity extraction (REE)} sub-task of template filling (Figure~\ref{fig:task}).\footnote{In this work, we assume there is one generic template for the entire document~\cite{huang-riloff-2011-peeling,huang2012modeling}.} 
The input text describes a \bombing \ event; the goal is to identify the entities that fill any of the roles associated with the event (e.g., the perpetrator, their organization, the weapon) by extracting a descriptive ``mention'' of it -- a string from the document. 

In contrast to sentence-level event extraction (see, e.g., the ACE evaluation \cite{ace-2005}), document-level REE introduces several complications.
First, \textbf{role-filler entities must be extracted even if they never appear in the same sentence as an event trigger}. In  Figure~\ref{fig:task}, for example, the \weapon \ and the first mention of the telephone company building (\target) \ appear in a sentence that does not explicitly mention the explosion of the bomb. 
In addition, \textbf{REE is ultimately an entity-based task} --- exactly one descriptive mention for each role-filler should be extracted even when the entity is referenced multiple times in connection with the event. The final output for the bombing example should, therefore, include just one of the ``water pipes'' references, and one of the three alternative descriptions of the \perpind \ and the second \target, the telephone company building. 
%
%
As a result of these complications, end-to-end sentence-level event extraction models~\cite{chen-etal-2015-event, lample-etal-2016-neural}, which dominate the literature, are ill-suited for the REE task, which calls for models that encode information and track entities across a longer context.
%
%
%
%
 
Fortunately, neural models for event extraction that have the ability to model longer contexts have been developed. 
\newcite{du2020document}, for example, extend standard  contextualized representations~\cite{devlin-etal-2019-bert} to produce a document-level sequence tagging model for event argument extraction.  Both approaches show improvements in performance over sentence-level models on event extraction.
Regrettably, these approaches (as well as most sentence-level 
methods) handle each candidate role-filler prediction in isolation.  Consequently, they cannot easily \textbf{model the coreference structure required to limit spurious role-filler mention extractions}. Nor can they easily \textbf{exploit semantic dependencies between closely related roles} like the \perpind \ and the \perporg, which 
can share a portion of the same entity span.  ``Shining Path 
members'', for instance, describes the \perpind \ in 
Figure~\ref{fig:task}, and its sub-phrase, ``Shining Path'', describes 
the associated \perporg.


\paragraph{Contributions} 
In this work we revisit the classic but recently under-studied  problem of document-level role-filler entity extraction problem and introduce a novel end-to-end generative transformer model --- the  ``\textbf{G}enerative \textbf{R}ole-f\textbf{i}ller \textbf{T}ransformer'' (\textbf{GRIT}) (Figure~\ref{fig:model}).
\begin{itemize}[leftmargin=*]
    \item Designed to model context at the document level, GRIT (1) has the ability to make extraction decisions across sentence boundaries; (2) is implicitly aware of noun phrase coreference structure;
    and (3) has the capacity to respect cross-role dependencies. More specifically, GRIT is built upon the pre-trained transformer model (BERT): we add a pointer selection module in the decoder to permit access to the entire input document, and a generative head to model document-level extraction decisions.
In spite of the added extraction capability, GRIT requires no additional parameters beyond those in the pre-trained BERT.
    
    \item To measure the model's ability to both extract entities for each role, and implicitly recognize coreferent relations between entity mentions, we design a metric (CEAF-REE) based on a maximum bipartite matching algorithm, drawing insights from the CEAF~\cite{luo-2005-coreference} coreference resolution measure.
    
    \item We evaluate GRIT on 
    the MUC-4~\shortcite{muc-1992-message} REE task (Section~\ref{sec:task_definition}). Empirically, our model outperforms substantially strong baseline models. 
    We also demonstrate that GRIT is better than existing document-level event extraction approaches at capturing linguistic properties critical for the task, including coreference between entity mentions and cross-role extraction dependencies.\footnote{Our code for the evaluation script and models is at \url{https://github.com/xinyadu/grit_doc_event_entity} for reproduction purposes.}
\end{itemize}


\section{Related Work}


\paragraph{Sentence-level Event Extraction}
Most work in event extraction has focused on the ACE sentence-level event task~\cite{walker2006ace},
which requires the detection of an event trigger and extraction of its arguments from within a single sentence. 
%
Previous state-of-the-art methods include \newcite{li-etal-2013-joint} and \newcite{li-etal-2015-improving-event}, which explored a variety of hand-designed features. 
More recently, neural network based models such as recurrent neural networks~\cite{nguyen-etal-2016-joint, feng2018language}, convolutional neural networks~\cite{nguyen-grishman-2015-event, chen-etal-2015-event} and attention mechanisms~\cite{liu-etal-2017-exploiting, liu-etal-2018-jointly} have also been shown to help improve performance.
Beyond the task-specific features learned by the deep neural models, \newcite{zhangtongtao2019joint} and \newcite{wadden-etal-2019-entity} also utilize pre-trained contextualized representations.

Only a few models have gone beyond individual sentences to make decisions.
\newcite{ji-grishman-2008-refining} and \newcite{liao-grishman-2010-using} utilize event type co-occurrence patterns to propagate event classification decisions.
\newcite{yang-mitchell-2016-joint} propose to learn within-event (sentence) structures for jointly extracting events and entities within a document context. 
Similarly, from a methodological perspective, our GRIT model also learns structured information, but it learns the dependencies between role-filler entity mentions and between different roles.
\newcite{duan-etal-2017-exploiting} and \newcite{zhao-etal-2018-document} leverage document embeddings as additional features to aid event detection.
Although the approaches above make decisions with cross-sentence information, their extractions are still done the sentence level.

\paragraph{Document-level IE}

Document-level event role-filler \textit{mention} extraction has been explored in recent work,   
using hand-designed features for both local and additional context~\cite{patwardhan-riloff-2009-unified, huang-riloff-2011-peeling, huang2012modeling},
and with end-to-end sequence tagging based models with contextualized pre-trained representations~\cite{du2020document}.
These efforts are the most related to our work. The key difference is that our work focuses on a more challenging, and more realistic, setting: extracting \textbf{role-filler entities} rather than lists of role-filler mentions that are not grouped according to their associated entity.
%
Also on a related note, \newcite{chambers-jurafsky-2011-template}, \newcite{chambers-2013-event}, and \newcite{liu-etal-2019-open} work on \textit{unsupervised} event schema induction and open-domain event extraction from documents. The main idea is to group entities corresponding to the same role into an event template. 

Recently, there has also been increasing interest in cross-sentence/document-level relation extraction (RE).
In the scientific domain, \newcite{peng-etal-2017-cross, wang-poon-2018-deep, jia-etal-2019-document} study $N$-ary cross-sentence RE using distant supervision annotations. \newcite{luan-etal-2018-multi} introduce SciERC dataset and their model rely on multi-task learning to share representations between entity span extraction and relations.
\newcite{yao2019docred} construct an RE dataset of cross-sentence relations on Wikipedia paragraphs.
\newcite{ebner-etal-2020-multi} introduce RAMS dataset for multi-sentence argument \textit{mention} linking, while we focus on entity-level extraction in our work.
Different from work on joint modeling~\cite{miwa-bansal-2016-end} and multi-task learning~\cite{luan-etal-2019-general} setting for extracting entities and relations, through the generative modeling setup, our GRIT model \textit{implicitly} captures (non-)coreference relations between noun phrases, without relying on the cross-sentence coreference and relation annotations during training.

\paragraph{Neural Generative Models with a Shared Module for Encoder and Decoder}
Our GRIT model uses one shared transformer module for both the encoder and decoder, which is simple and effective.
For the machine translation task, \newcite{he2018layer} propose a model which shares the parameters of each layer between the encoder and decoder to regularize and coordinate the learning.
\newcite{dong2019unified} presents a new unified pre-trained language model that can be fine-tuned for both NLU and NLG tasks. Similar to our work, they also introduce different masking strategies for different kinds of tasks (see Section\ref{sec:method}).

\section{The Role-filler Entity Extraction Task and Evaluation Metric}
\label{sec:task_definition}

We base the REE task on the original MUC\footnote{The Message Understanding Conferences were a series of U.S. government-organized IE evaluations.} formulation~\cite{sundheim-1991-overview}, but simplify it as done in prior  research~\cite{huang2012modeling, du2020document}.
In particular, we assume that 
one generic template should be produced for each document: for documents that recount more than one event, the extracted role-filler entities for each are merged into a single event template.
Second, we focus on entity-based roles with string-based fillers\footnote{Other types of role fillers include normalized dates and times, and categorical ``set" fills. We do not attempt to handle these in the current work.}.
%
\begin{itemize}[leftmargin=*]
    \item Each \textit{event} consists of the set of roles that describe it (shown in Figure~\ref{fig:task}). The MUC-4 dataset that we use consists of $\sim$1k terrorism events.
    
    \item Each \textit{role} is filled with one or more entities. There are five such roles for MUC-4: perpetrator individuals (\perpind), perpetrator organizations (\perporg), physical targets (\target), victims (\victim) and weapons (\weapon). These event roles represent the agents, patients, and instruments associated with terrorism events~\cite{huang2012modeling}.
    
    \item Each \textit{role-filler entity} is denoted by a single descriptive \textit{mention}, a span of text from the input document. Because multiple such mentions for each entity may appear in the input, the gold-standard template lists all alternatives (shown in Figure~\ref{fig:task}), but systems are required to produce just one.
    \end{itemize}

    
\paragraph{Evaluation Metric}
\label{sec:ceaf-REE}
The metric for past work on document-level role-filler \textit{mentions} extraction~\cite{patwardhan-riloff-2009-unified, huang-riloff-2011-peeling, du2020document} calculates mention-level precision across all alternative mentions for each role-filler entity. Thus it is not suited for our problem setting, where entity-level precision is needed, where spurious entity extractions will get punished (e.g., recognizing ``telephone company building'' and ``telephone company offices'' as two entities will result in lower precision).

Drawing insights from the entity-based CEAF metric~\cite{luo-2005-coreference} from the coreference resolution literature, we design a metric (\textbf{CEAF-REE}) for measuring models' performance on this document-level \textbf{role-filler entity extraction} task. 
It is based on maximum bipartite matching algorithm~\cite{kuhn1955hungarian,munkres1957algorithms}. 
The general idea is that, for each role, the metric is computed by aligning gold and predicted entities with the constraint that a predicted (gold) entity is aligned with at most one gold (predicted) entity. Thus, the system that does not recognize the coreferent mentions and use them for separate entities will be penalized in precision score. For the example in Figure~\ref{fig:task}, if the system extracts ``Pilmai telephone company building'' and ``telephone company offices'' as two distinct \target s, the precision will drop. We include more details for our CEAF-TF metric in the appendix.

\section{REE as Sequence Generation}






We treat document-level REE as a sequence-to-sequence task \cite{sutskever2014sequence} in order to better model the cross-role dependencies and cross-sentence noun phrase coreference structure. We first transform the task definition into a source and target sequence. 

\begin{figure*}[t]
\centering
\resizebox{\textwidth}{!}{
\includegraphics[]{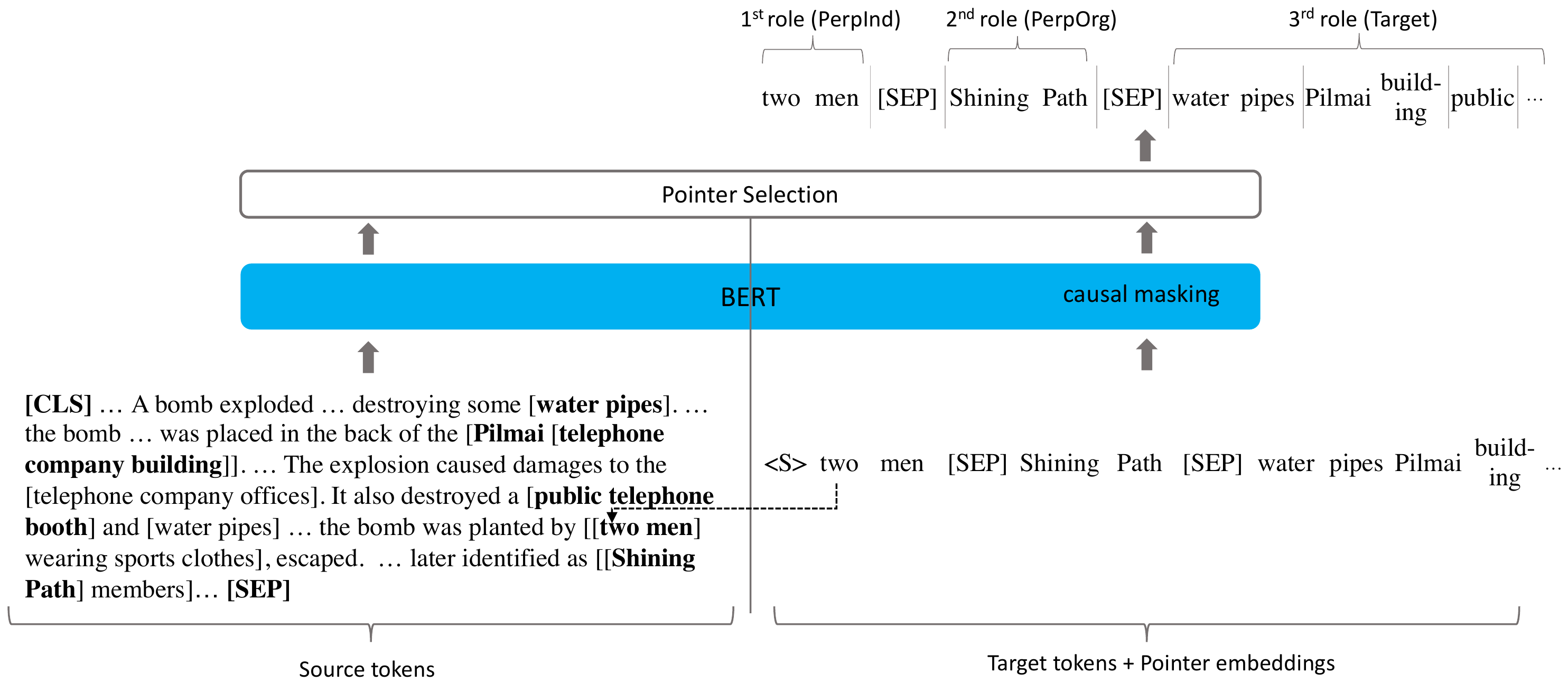}
}
\caption{GRIT: generative transformer model for document-level event role-filler entity extraction. (Noun phrase bracketing and \textbf{bold} in the source tokens are provided for readability purposes and are not part of the source sequence.)}
\label{fig:model}
\end{figure*}

As shown in Figure~\ref{fig:model}, the \textit{source sequence} simply consists of the tokens of the original document prepended with a ``classification'' token (i.e., [CLS] in BERT), and appended with a separator token (i.e., [SEP] in BERT).
%
The \textit{target sequence} is the concatenation of target extractions for each role, separated by the separator token.
For each role, the target extraction consists of the first mention's beginning ($b$) and end ($e$) tokens:
\begin{equation}
\begin{gathered}
\nonumber
\text{<S>} \ e^{(1)}_{1_b}, e^{(1)}_{1_e},... \ \text{[SEP]} \\ 
\ \ e^{(2)}_{1_b}, e^{(2)}_{1_e},... \ \text{[SEP]} \\
\quad e^{(3)}_{1_b}, e^{(3)}_{1_e}, e^{(3)}_{2_b}, e^{(3)}_{2_e}, ... \ \text{[SEP]} \\
... \ 
\end{gathered}
\end{equation}
Note that we list the roles in a fixed order for all examples.  So for the example used in Figure~\ref{fig:model}, $e^{(1)}_{1_b}$, $e^{(1)}_{1_e}$ would be ``two'' and ``men'' respectively; and $e^{(3)}_{1_b}$, $e^{(3)}_{1_e}$ would be ``water'' and ``pipes'' respectively.
Henceforth, we denote the resulting sequence of source tokens as $x_0, x_1, ..., x_m$ and the sequence of target tokens as $y_0, y_1, ..., y_n$.


\section{Model: Generative Role-filler Transformer (GRIT)}
\label{sec:method}

Our model is shown in Figure~\ref{fig:model}. It consists of two parts: the encoder (left) for the source tokens; and the decoder (right) for the target tokens. Instead of using a sequence-to-sequence learning architecture with separate modules~\cite{sutskever2014sequence, bahdanau2014neural}, we use a single pretrained transformer model~\cite{devlin-etal-2019-bert} for both parts, and introduce no additional fine-tuned parameters. 



\begin{figure}[t]
\resizebox{\columnwidth}{!}{
\includegraphics{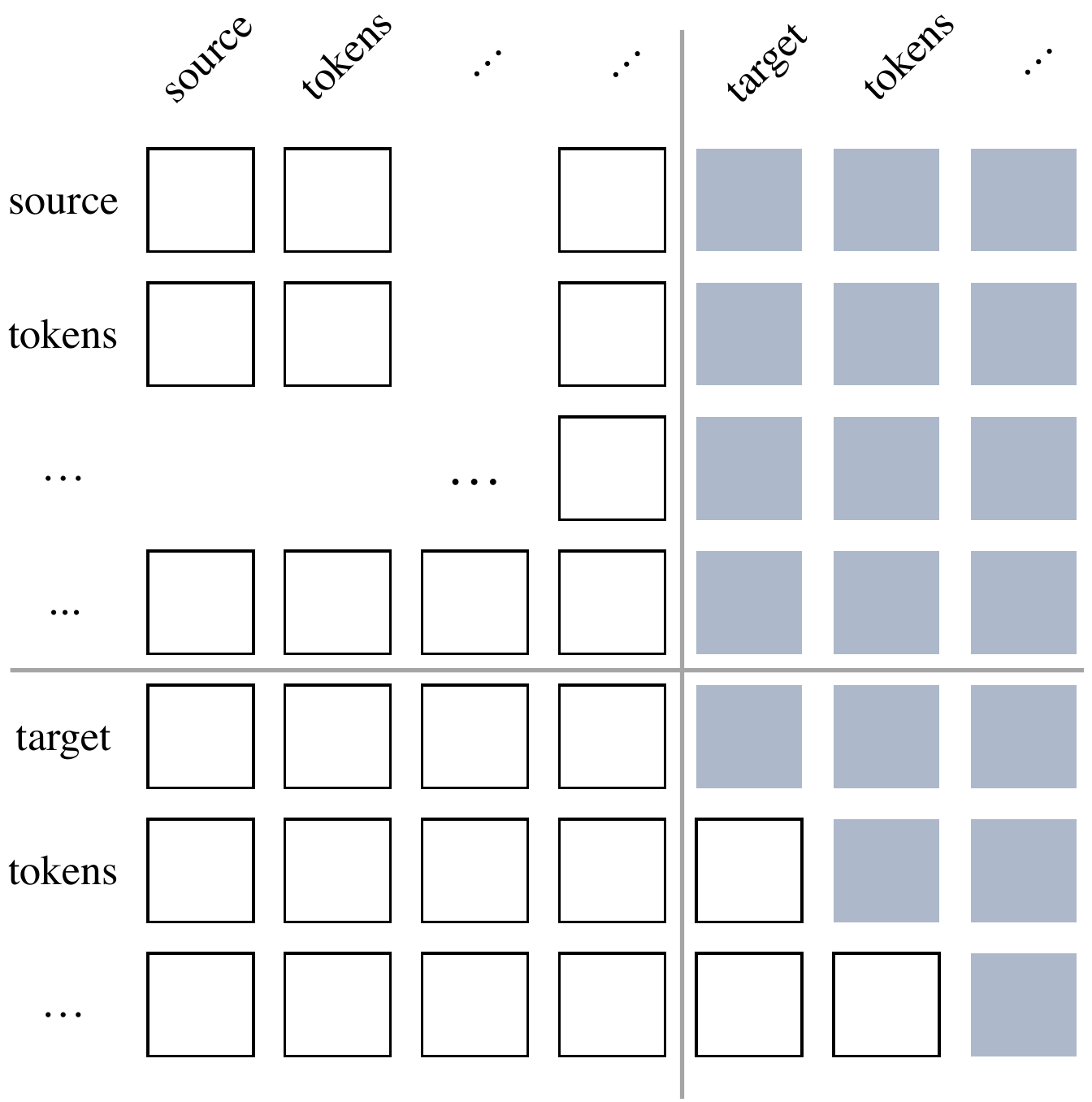}
}
\caption{Partially causal masking strategy ($\mathbf{M}$). (White cell: unmasked; Grey cell: masked).}
\label{fig:masking}
\end{figure}

\paragraph{Pointer Embeddings}


The first change to the model is to ensure that the decoder is aware of where its previous predictions come from in the source document, an  approach we call ``pointer embeddings''.  Similar to BERT, the input to the model consists of the sum of token, position and segment embeddings. However, for the position we use the corresponding source token's position. For example, for the word ``two'', the target tokens would have the identical position embedding of the word ``two'' in the source document. Interestingly, we do not use any explicit target position embeddings, but instead separate each role with a [SEP] token. Empirically, we find that the model is able to use these separators to learn which role to fill and which mentions have filled previous roles. 

Our encoder's embedding layer uses standard BERT embedding layer, which applied to the source document tokens. To denote boundary between source and target tokens, we use sequence A (first sequence) segment embeddings for the source tokens, we use sequence B (second sequence) segment embeddings for the target tokens.

We pass the source document tokens through the encoder's embedding layer, to obtain their embeddings $\mathbf{x}_{0}, \mathbf{x}_{1}, ..., \mathbf{x}_{m}$.
We pass the target tokens $y_0, y_1, ..., y_n$ through the decoder's embedding layer, to obtain $\mathbf{y}_{0}, \mathbf{y}_{1}, ..., \mathbf{y}_{n}$.

\paragraph{BERT as Encoder / Decoder}
We utilize one BERT model as both the source and target embeddings. To distinguish the encoder / decoder representations, we provide a partial causal attention mask on the decoder side. 

In Figure~\ref{fig:masking}, we provide an illustration for the attention masks -- 2-dimensional matrix denoted as $\mathbf{m}$. 
For the source tokens, the mask allows full source self-attention,  but mask out all target tokens. For $i \in \{0,1,...,m\}$,
\begin{equation}
\nonumber
\mathbf{M}_{i,j}=
\left\{
\begin{aligned}
1 &, \ \ \text{if} \ \ 0 \leq j \leq m \\
0 &, \ \ \text{otherwise} \\
\end{aligned}
\right.
\end{equation}

For the target tokens, to guarantee that the decoder is autoregressive (the current token should not attend to future tokens), we use a causal masking strategy. Assuming 
we concatenate the target to the source tokens (the joint sequence mentioned below), for $i \in \{m+1,...,n\}$,
\begin{equation}
\nonumber
\mathbf{M}_{i,j}=
\left\{
\begin{aligned}
1 &, \ \ \text{if} \ \ 0 \leq j \leq m \\
1 &, \ \ \text{if} \ \ j > m \ \ \text{and} \ \ j \leq i \\
0 &, \ \ \text{otherwise} \\
\end{aligned}
\right.
\end{equation}

The joint sequence of source tokens' embeddings ($\mathbf{x}_0, \mathbf{x}_1, ..., \mathbf{x}_{m}$) and target tokens' embeddings ($\mathbf{y}_{0}, \mathbf{y}_{1}, ..., \mathbf{y}_{n}$) are passed through BERT to obtain their contextualized representations,
\begin{equation}
\begin{gathered}
\nonumber
\mathbf{\hat{x}}_0, \mathbf{\hat{x}}_1, ..., \mathbf{\hat{x}}_{m}, \mathbf{\hat{y}}_{0}..., \mathbf{\hat{y}}_{n} \\
= \verb|BERT| (\mathbf{x}_0, \mathbf{x}_1, ..., \mathbf{x}_{m}, \mathbf{y}_{0}, ..., \mathbf{y}_{n})
\end{gathered}
\end{equation}


\paragraph{Pointer Decoding}

For the final layer, we replace word prediction with a simple pointer selection mechanism. For target time step $t \ (0 \leq t \leq n)$, we first calculate the dot-product between $\mathbf{\hat{y}}_t$ and 
$\mathbf{\hat{x}}_0, \mathbf{\hat{x}}_1, ..., \mathbf{\hat{x}}_{m}$,
\begin{equation}
\nonumber
z_0, z_1, ..., z_m = \mathbf{\hat{y}}_t \cdot \mathbf{\hat{x}}_0, \mathbf{\hat{y}}_t \cdot \mathbf{\hat{x}}_1, ..., \mathbf{\hat{y}}_t \cdot \mathbf{\hat{x}}_m
\end{equation}
Then we apply softmax to $z_0, z_1, ..., z_m$ to obtain the probabilities of pointing to each source token,
\begin{equation}
\nonumber
p_0, p_1, ..., p_m = \text{softmax}(z_0, z_1, ..., z_m)
\end{equation}


Test prediction is done with greedy decoding. 
At each time step $t$, $\text{argmax}$ is applied to find the source token which has the highest probability. The predicted token is added to the target sequence for the next time step $t+1$ with its pointer embedding. We stop decoding when the fifth [SEP] token is predicted, which represents the end of extractions for the last role.

In addition, we add the following decoding constraints,
\begin{itemize}[leftmargin=*] 
    \item Tune probability of generating [SEP]. By doing this, we encourage the model to point to other source tokens and thus extract more entities for each role, which will help increase the recall.  (We set the hyperparameter of downweigh to $0.01$, i.e.,
    for the [SEP] token $p = 0.01 * p$.)
    \item Ensure that the token position increase from start token to end token. When decoding tokens for each role, we know that mention spans should obey this property. Thus we eliminate those invalid choices during decoding.
\end{itemize}

\section{Experimental Setup}

We conduct evaluations on the MUC-4 dataset~\shortcite{muc-1992-message}, and compare to recent competitive end-to-end models~\cite{wadden-etal-2019-entity, du2020document} in IE (Section~\ref{sec:results}).
Besides the normal evaluation, we are also interested in how well our GRIT model captures coreference linguistic knowledge, and comparison with the prior models. In Section~\ref{sec:linguistic}, we present relevant evaluations on the subset of test documents.

\paragraph{Dataset and Evaluation Metric}

The MUC-4 dataset consists of 1,700 documents with associated templates. Similar to~\cite{huang2012modeling, du2020document}, we use the 1300 documents for training, 200 documents (\verb|TST1+TST2|) as the development set and 200 documents (\verb|TST3+TST4|) as the test set.
Each document in the dataset contains on average 403.27 tokens, 7.12 paragraphs. In Table~\ref{tab:role_des}, we include descriptions for each role in the template.

\begin{table}[t]
\resizebox{\columnwidth}{!}{
\begin{tabular}{l|l}
\toprule
Roles & Descriptions \\ \midrule
\perpind & A person responsible for the incident. \\ 
\perporg & An organization responsible for the incident. \\
\target  & A thing (inanimate object) that was attacked. \\
\begin{tabular}[c]{@{}l@{}}\victim \\ \ \\ \ \end{tabular}  & \begin{tabular}[c]{@{}l@{}}The name of a person who was the obvious \\ or apparent target of  the attack \\ or who became a victim of the attack.\end{tabular} \\
\weapon  & A device used by the perpetrator(s) in carrying. \\
\bottomrule
\end{tabular}}
\caption{Natural Language Descriptions for Each Role.}
\label{tab:role_des}
\end{table}

We use the first appearing mention of the role-filler entity as the training signal (thus do not use the other alternative mentions during training). 
    
%
We use CEAF-REE which is covered in Section~\ref{sec:ceaf-REE} as the evaluation metric.
The results are reported as Precision (P), Recall (R) and F-measure (F1) score for the micro-average for all the event roles (Table~\ref{tab:micro}). We also report the per-role results to have a fine-grained understanding of the numbers (Table~\ref{tab:per_role}).

\begin{table*}[t]
\resizebox{\textwidth}{!}{
\begin{tabular}{l|c|c|c|c|c}
\toprule
& \multicolumn{1}{c|}{\perpind} & \multicolumn{1}{c|}{\perporg} & \multicolumn{1}{c|}{\target} & \multicolumn{1}{c|}{\victim} & \multicolumn{1}{c}{\weapon} \\ \midrule
\begin{tabular}[c]{@{}l@{}}NST \\ \cite{du2020document} \end{tabular} & 48.39 / 32.61 / 38.96   & 60.00 / 43.90 / 50.70   & 54.96 / 52.94 / \textbf{53.93}  & 62.50 / 63.16 / 62.83  & 61.67 / 61.67 / \textbf{61.67}  \\
\begin{tabular}[c]{@{}l@{}}\dygie \\ \cite{wadden-etal-2019-entity} \end{tabular}  & 59.49 / 34.06 / 43.32   & 56.00 / 34.15 / 42.42   & 53.49 / 50.74 / 52.08  & 60.00 / 66.32 / 63.00  & 57.14 / 53.33 / 55.17  \\ \midrule
GRIT & 65.48 / 39.86 / \textbf{49.55}   & 66.04 / 42.68 / \textbf{51.85}   & 55.05 / 44.12 / 48.98  & 76.32 / 61.05 / \textbf{67.84}  & 61.82 / 56.67 / 59.13 \\ \bottomrule
\end{tabular}}
\caption{Per-role performance scored by CEAF-REE (reported as P/R/F1, highest F1 for each role are boldfaced).}
\label{tab:per_role}
\end{table*}

\begin{table*}[t]
\resizebox{\textwidth}{!}{
\begin{tabular}{l|c|c|c|c|c}
\toprule
& \multicolumn{1}{c|}{$k=1$} & \multicolumn{1}{c|}{$1< k \leq 1.25$} & \multicolumn{1}{c|}{$1.25 < k \leq 1.5$} & \multicolumn{1}{c|}{$1.5 < k \leq 1.75$} & \multicolumn{1}{c}{$k > 1.75$} \\ \midrule
\begin{tabular}[c]{@{}l@{}}NST \\\cite{du2020document}\end{tabular} & 63.83 / 51.72 / 57.14 & 57.45 / 38.57 / 46.15          & 60.32 / 49.03 / 54.09           & 64.81 / 50.00 / 56.45          & 66.67 / 51.90 / 58.36 \\
\begin{tabular}[c]{@{}l@{}}\dygie \\ \cite{wadden-etal-2019-entity} \end{tabular}  & \textbf{72.50} / 50.00 / 59.18 & 70.00 / 40.00 / 50.91          & 60.48 / 48.39 / 53.76   & 52.94 / 38.57 / 44.63  & 66.96 / 48.73 / 56.41\\ \midrule
GRIT & 65.85  / 46.55 / 54.55 & \textbf{74.42} / 45.71 / 56.64  & \textbf{73.20} / 45.81 / 56.35 & \textbf{67.44} / 41.43 / 51.33 & \textbf{69.75} / 52.53 / 59.93    \\ \bottomrule
\end{tabular}}
\caption{Evaluations on the subsets of documents with increasing number of mentions per role-filler entity. $k$ denotes the average \# mentions per role-filler entity. Results for each column are reported as Precision / Recall / F1. The highest precisions are boldfaced for each bucket.}
\label{tab:inter_entity}
\end{table*}

\paragraph{Baselines}

We compare to recent strong models for (document-level) information/event extraction.
CohesionExtract~\cite{huang2012modeling} is a bottom-up approach for event extraction that first aggressively identifies candidate role-fillers, and prune the candidates located in event-irrelevant sentences.\footnote{Instead of using feature-engineering based sentence classification to identify event-relevant sentences, we re-implement the sentence classifier with BiLSTM-based neural sequence model.}
%
\newcite{du2020document} propose \textit{neural sequence tagging (NST)} models with contextualized representations for document-level role filler mentions extraction. We train this model with BIO tagging scheme to identify the first mention for each role-filler entity and its type (i.e., B-PerpInd, I-PerpInd for perpetrator individual).
\textit{\dygie}~\cite{wadden-etal-2019-entity} is a span-enumeration based extraction model for entity, relation, and event extraction. 
The model 
(1) enumerates all the possible spans in the document;
(2) concatenates the representations of the span's beginning \& end token and use it as its representation, and pass it through a classifier layer to predict whether the span represents certain role-filler entity and what the role is.
Both the NST and \dygie \ are end-to-end and fine-tuned BERT~\cite{devlin-etal-2019-bert} contextualized representations with task-specific data. We train them to identify the first mention for each role-filler entity (to ensure fair comparison with our proposed model).
Unsupervised event schema induction based approaches~\cite{chambers-jurafsky-2011-template, chambers-2013-event, cheung-etal-2013-probabilistic} are also able to model the coreference relations and entities at document-level, but have been proved to perform substantially worse than supervised models~\cite{patwardhan-riloff-2009-unified, huang2012modeling}. Thus we do not compare with them.
We also experimented with a variant of our GRIT model -- instead of always pointing to the same [SEP] in the source tokens to finish extracting the role-filler entities for a role, we use five different [SEP] tokens. During decoding, the model points to the corresponding [SEP] as the end of extraction for that role. This variant does not improve over the current best results and we omit reporting its performance.

\section{Results}
\label{sec:results}

In Table~\ref{tab:micro}, we report the micro-average performance on the test set. We observe that our GRIT model substantially outperforms the baseline extraction models in precision and F1, with an over 5\% improvement in precision over \dygie.

Table~\ref{tab:per_role} compares the models' performance scores on each role (\perpind, \perporg, \target, \victim, \weapon). 
We see that, 
(1) our model achieves the best precision across the roles; 
(2) for the roles that come with entities containing more human names (e.g., \perpind \ and \victim), our model substantially outperforms the baselines; 
(3) for the role \perporg, our model scores better precision but lower recall than {neural sequence tagging}, which results in a slightly better F1 score;
(4) for the roles \target \ and \weapon, our model is more conservative (lower recall) and achieves lower F1. One possibility is that for role like \target, on average there are more entities (though with only one mention each), and it's harder for our model to decode as many \target \ entities correct in a generative way.

\begin{table}[t]
\centering
\resizebox{\columnwidth}{!}{
\begin{tabular}{l|ccc}
\toprule
Models      & P     & R     & F1    \\ \midrule
\begin{tabular}[c]{@{}l@{}}CohesionExtract \\ \cite{huang2012modeling} \end{tabular} & 58.38 & 39.53 & 47.14 \\
\begin{tabular}[c]{@{}l@{}}NST \\ \cite{du2020document} \end{tabular} & 56.82 & \textbf{48.92} & 52.58 \\
\begin{tabular}[c]{@{}l@{}}\dygie \\ \cite{wadden-etal-2019-entity}  \end{tabular}     & 57.04 & 46.77 & 51.40 \\ \midrule
GRIT     & \textbf{64.19}$^{**}$ & 47.36 & \textbf{54.50}$^{*}$ \\
\bottomrule
\end{tabular}
}
\caption{Micro-average results (the highest number of each column is boldfaced). \small {Significance is indicated with $^{**}$($p<0.01$),$^*$($p<0.1$) -- all tests are computed using the paired bootstrap procedure~\cite{berg-kirkpatrick-etal-2012-empirical}.}
}
\label{tab:micro}
\end{table}

\section{Discussion}

\paragraph{How well do the models capture coreference relations between mentions?}
\label{sec:linguistic}

We also conduct targeted evaluations on subsets of test documents whose gold extractions come with coreferent mentions. From left to right in Table~\ref{tab:inter_entity}, we report results on the subsets of documents with increasing number ($k$) of possible (coreferent) mentions per role-filler entity. We find that: 
(1) On the subset of documents with only one mention for each role-filler entity ($k=1$), our model has no significant advantage over \dygie \ and the sequence tagging based model;
(2) But as $k$ increases, the advantage of our GRIT substantially increases -- with an over 10\% gap in precision when $1<k\leq1.5$, and a near 5\% gap in precision when $k > 1.5$.

\begin{figure}[h]
\centering
\resizebox{\columnwidth}{!}{
\includegraphics[]{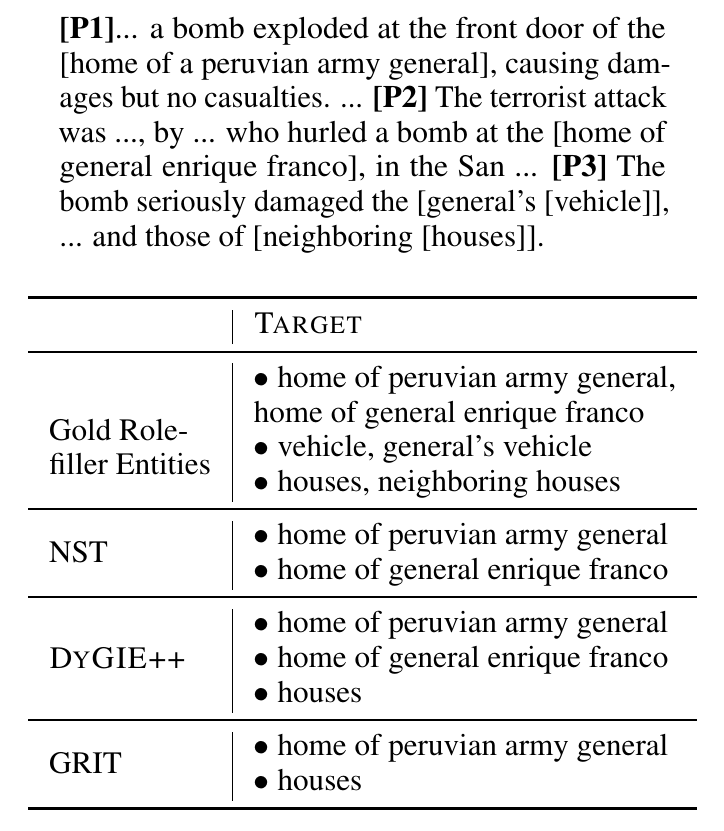}
}
\caption{Our model implicitly captures coreference relations between mentions.}
\label{fig:coref_e}
\end{figure}

From the qualitative example (document excerpt and the extractions in Figure~\ref{fig:coref_e}), 
we also observe our model recognizes the coreference relation between candidate role-filler entity mentions, while the baselines do not, which shows that our model is better at capturing the (non-)coreference relations between role-filler entity mentions. It also proves the advantage of a generative model in this setting.

\paragraph{How well do models capture dependencies between different roles?}

To study this phenomenon, we consider nested role-filler entity mentions in the documents. In the example of Figure~\ref{fig:task}, {``shining path''} is a role-filler entity mention for \perporg \  nested in {``two \textit{shining path} members''} (a role-filler entity mention for \perpind). The nesting happens more often between more related roles (e.g., \perpind \ and \perporg) --  we find that 33 out of the 200 test documents' gold extractions contain nested role-filler entity mentions between the two roles.

\begin{table}[ht]
\centering
\resizebox{\columnwidth}{!}{
\begin{tabular}{l|c|c}
\toprule
    & \multicolumn{1}{c|}{\perporg \ (all docs)} & \multicolumn{1}{c}{\perporg \ (33/200)} \\ \midrule
& P / R / F1           & P / R / F1          \\ \midrule
\begin{tabular}[c]{@{}l@{}}NST \end{tabular} & 56.00  / 34.15 / 42.42        & 80.00  / 44.44 / 57.14       \\
\begin{tabular}[c]{@{}l@{}}\dygie  \end{tabular} & 60.00  / \textbf{43.90} / 50.70        & 61.54 / 35.56 / 45.07 \\ \midrule
GRIT & 66.04 / 42.68 / 51.85        & 80.77 / \textbf{46.67} / 59.15      \\ \bottomrule
\end{tabular}}
\caption{Evaluation on the subset of documents that have nested role-filler entity mentions between role \perpind \ and \perporg \ (highest recalls boldfaced).}
\label{tab:inter_role}
\end{table}

\begin{figure}[h]
\centering
\resizebox{\columnwidth}{!}{
\includegraphics[]{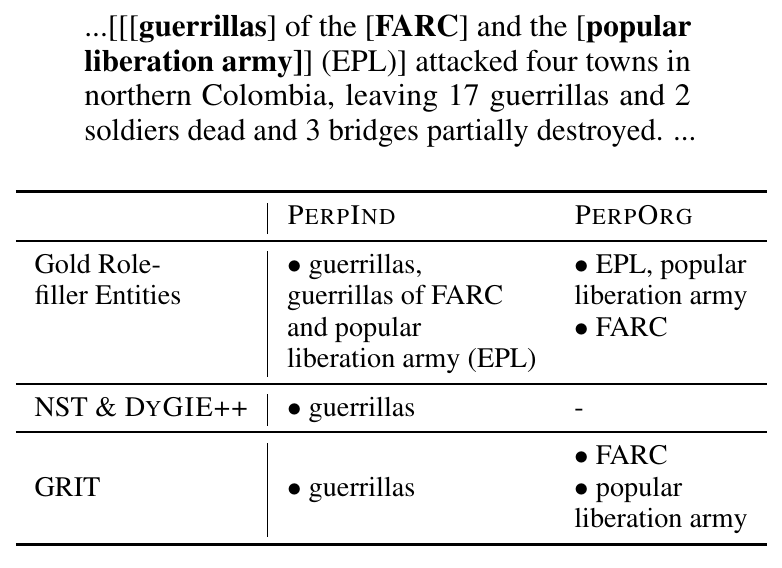}
}
\caption{Our model captures dependencies between different roles.}
\label{fig:role_e}
\end{figure}

In Table~\ref{tab:inter_role}, we present the CEAF-REE scores for role \perporg \ on the subset of documents with nested roles. As we hypothesized beforehand, GRIT is able to learn the dependency between different roles and can learn to avoid missing relevant role-filler entities for later roles.
The results provide empirical evidence: by learning the dependency between \perpind \ and \perporg, GRIT improves the relative recall score on the subset of documents as compared to \dygie. On all the 200 test documents, our model is $\sim2\%$ below \dygie \ in recall; while on the 33 docs, our model scores much higher than \dygie \ in recall. 

For the document in the example of Figure~\ref{fig:role_e}, our model correctly extracts the two role-filler entities for \perporg: ``FARC'' and ``popular liberation army'', which are closely related to the \perpind \ entity ``guerrilla''. While \dygie \  and NST both miss the entities for \perporg.

\paragraph{Decoding Ablation Study}

In the table below, we present ablation results based on the decoding constraints. These illustrate the influence of the decoding constraints on the our model's performance. The two constraints both significantly improve model predictions. 
Without downweighing the probability of pointing to [SEP], the precision increases but recall and F1 significantly drops.

\begin{table}[h]
\small
\centering
\resizebox{\columnwidth}{!}{
\begin{tabular}{l|ccc|l}
\toprule
& P & R & F1 & $\Delta$ (F1) \\ \midrule
GRIT        &  64.19 & 47.36 & 54.50 &         \\
$-$ {[}SEP{]} downweigh  &  67.43 & 40.12 & 50.31 & -4.19 \\
$-$ constraint on pointer offset  &     62.90 & 45.79 & 53.00 & -1.50    \\
\bottomrule
\end{tabular}}
\caption{Decoding Ablation Study}
\label{tab:ablation}
\end{table}


\paragraph{Additional Parameters and Training Cost} 

Finally we consider additional parameters and training time of the models:
As we introduced previously, the baseline models \dygie \ and NST both require an additional classifier layer on top of BERT's hidden state (of size $H$) for making the predictions. While our GRIT model does not require adding any new parameters. As for the training time, training the \dygie \ model takes over 10 times longer time than NST and our model. This time comes from the  \dygie \ model requirement of enumerating all possible spans (to a certain length constraint) in the document and calculating the loss with their labels.

\begin{table}[h]
\label{tab:cost}
\small
\centering
\begin{tabular}{l|c|c}
\toprule
& additional params & training cost \\ \midrule
\begin{tabular}[c]{@{}l@{}}\dygie \\  \end{tabular}                   
& $2H(\# \text{roles}+1)$ & $\sim$20h          \\
\begin{tabular}[c]{@{}l@{}}NST \\  \end{tabular} 
& $H(2\# \text{roles}+1)$ & $\sim$1h           \\ \midrule
GRIT                 & 0                                         & \textless{}40min  \\ \bottomrule
\end{tabular}
\caption{Additional Parameters and Training Cost.}
\end{table}


\section{Conclusion} 
We revisit the classic and challenging problem of document-level role-filler entity extraction (REE), and find that there is still room for improvement. We introduce an effective end-to-end transformer based generative model, which learns the document representation and encodes the dependency between role-filler entities and between event roles. It outperforms the baselines on the task and better captures the coreference linguistic phenomena. In the future, it would be interesting
to investigate how to enable the model to also do template recognition.

\section*{Acknowledgments}
We thank the anonymous reviewers for helpful feedback and suggestions.
The work of XD and AMR was supported by NSF CAREER 2037519; that of XD and CC was supported in part by DARPA LwLL Grant FA8750-19-2-0039.

\bibliography{eacl2021}

\begin{thebibliography}{48}
\expandafter\ifx\csname natexlab\endcsname\relax\def\natexlab#1{#1}\fi

\bibitem[{Bahdanau et~al.(2015)Bahdanau, Cho, and Bengio}]{bahdanau2014neural}
Dzmitry Bahdanau, Kyunghyun Cho, and Yoshua Bengio. 2015.
\newblock \href {http://arxiv.org/abs/1409.0473} {Neural machine translation by
  jointly learning to align and translate}.
\newblock In \emph{3rd International Conference on Learning Representations,
  {ICLR} 2015, San Diego, CA, USA, May 7-9, 2015, Conference Track
  Proceedings}.

\bibitem[{Berg-Kirkpatrick et~al.(2012)Berg-Kirkpatrick, Burkett, and
  Klein}]{berg-kirkpatrick-etal-2012-empirical}
Taylor Berg-Kirkpatrick, David Burkett, and Dan Klein. 2012.
\newblock \href {https://www.aclweb.org/anthology/D12-1091} {An empirical
  investigation of statistical significance in {NLP}}.
\newblock In \emph{Proceedings of the 2012 Joint Conference on Empirical
  Methods in Natural Language Processing and Computational Natural Language
  Learning}, pages 995--1005, Jeju Island, Korea. Association for Computational
  Linguistics.

\bibitem[{Chambers(2013)}]{chambers-2013-event}
Nathanael Chambers. 2013.
\newblock \href {https://www.aclweb.org/anthology/D13-1185} {Event schema
  induction with a probabilistic entity-driven model}.
\newblock In \emph{Proceedings of the 2013 Conference on Empirical Methods in
  Natural Language Processing}, pages 1797--1807, Seattle, Washington, USA.
  Association for Computational Linguistics.

\bibitem[{Chambers and Jurafsky(2011)}]{chambers-jurafsky-2011-template}
Nathanael Chambers and Dan Jurafsky. 2011.
\newblock \href {https://www.aclweb.org/anthology/P11-1098} {Template-based
  information extraction without the templates}.
\newblock In \emph{Proceedings of the 49th Annual Meeting of the Association
  for Computational Linguistics: Human Language Technologies}, pages 976--986,
  Portland, Oregon, USA. Association for Computational Linguistics.

\bibitem[{Chen et~al.(2015)Chen, Xu, Liu, Zeng, and
  Zhao}]{chen-etal-2015-event}
Yubo Chen, Liheng Xu, Kang Liu, Daojian Zeng, and Jun Zhao. 2015.
\newblock \href {https://doi.org/10.3115/v1/P15-1017} {Event extraction via
  dynamic multi-pooling convolutional neural networks}.
\newblock In \emph{Proceedings of the 53rd Annual Meeting of the Association
  for Computational Linguistics and the 7th International Joint Conference on
  Natural Language Processing}, pages 167--176, Beijing, China. Association for
  Computational Linguistics.

\bibitem[{Cheung et~al.(2013)Cheung, Poon, and
  Vanderwende}]{cheung-etal-2013-probabilistic}
Jackie Chi~Kit Cheung, Hoifung Poon, and Lucy Vanderwende. 2013.
\newblock \href {https://www.aclweb.org/anthology/N13-1104} {Probabilistic
  frame induction}.
\newblock In \emph{Proceedings of the 2013 Conference of the North {A}merican
  Chapter of the Association for Computational Linguistics: Human Language
  Technologies}, pages 837--846, Atlanta, Georgia. Association for
  Computational Linguistics.

\bibitem[{Devlin et~al.(2019)Devlin, Chang, Lee, and
  Toutanova}]{devlin-etal-2019-bert}
Jacob Devlin, Ming-Wei Chang, Kenton Lee, and Kristina Toutanova. 2019.
\newblock \href {https://doi.org/10.18653/v1/N19-1423} {{BERT}: Pre-training of
  deep bidirectional transformers for language understanding}.
\newblock In \emph{Proceedings of the 2019 Conference of the North {A}merican
  Chapter of the Association for Computational Linguistics: Human Language
  Technologies}, pages 4171--4186, Minneapolis, Minnesota. Association for
  Computational Linguistics.

\bibitem[{Dong et~al.(2019)Dong, Yang, Wang, Wei, Liu, Wang, Gao, Zhou, and
  Hon}]{dong2019unified}
Li~Dong, Nan Yang, Wenhui Wang, Furu Wei, Xiaodong Liu, Yu~Wang, Jianfeng Gao,
  Ming Zhou, and Hsiao-Wuen Hon. 2019.
\newblock Unified language model pre-training for natural language
  understanding and generation.
\newblock In \emph{Advances in Neural Information Processing Systems}, pages
  13042--13054.

\bibitem[{Du and Cardie(2020)}]{du2020document}
Xinya Du and Claire Cardie. 2020.
\newblock \href {https://doi.org/10.18653/v1/2020.acl-main.714} {Document-level
  event role filler extraction using multi-granularity contextualized
  encoding}.
\newblock In \emph{Proceedings of the 58th Annual Meeting of the Association
  for Computational Linguistics}, pages 8010--8020, Online. Association for
  Computational Linguistics.

\bibitem[{Duan et~al.(2017)Duan, He, and Zhao}]{duan-etal-2017-exploiting}
Shaoyang Duan, Ruifang He, and Wenli Zhao. 2017.
\newblock \href {https://www.aclweb.org/anthology/I17-1036} {Exploiting
  document level information to improve event detection via recurrent neural
  networks}.
\newblock In \emph{Proceedings of the Eighth International Joint Conference on
  Natural Language Processing}, pages 352--361, Taipei, Taiwan. Asian
  Federation of Natural Language Processing.

\bibitem[{Ebner et~al.(2020)Ebner, Xia, Culkin, Rawlins, and
  Van~Durme}]{ebner-etal-2020-multi}
Seth Ebner, Patrick Xia, Ryan Culkin, Kyle Rawlins, and Benjamin Van~Durme.
  2020.
\newblock \href {https://doi.org/10.18653/v1/2020.acl-main.718} {Multi-sentence
  argument linking}.
\newblock In \emph{Proceedings of the 58th Annual Meeting of the Association
  for Computational Linguistics}, pages 8057--8077, Online. Association for
  Computational Linguistics.

\bibitem[{Feng et~al.(2018)Feng, Qin, and Liu}]{feng2018language}
Xiaocheng Feng, Bing Qin, and Ting Liu. 2018.
\newblock A language-independent neural network for event detection.
\newblock \emph{Science China Information Sciences}, 61(9):092106.

\bibitem[{Grishman and Sundheim(1996)}]{grishman-sundheim-1996-design}
Ralph Grishman and Beth Sundheim. 1996.
\newblock \href {https://doi.org/10.3115/1119018.1119072} {Design of the
  {MUC}-6 evaluation}.
\newblock In \emph{TIPSTER TEXT PROGRAM PHASE II: Proceedings of a Workshop
  held at Vienna, Virginia, May 6-8, 1996}, pages 413--422, Vienna, Virginia,
  USA. Association for Computational Linguistics.

\bibitem[{He et~al.(2018)He, Tan, Xia, He, Qin, Chen, and Liu}]{he2018layer}
Tianyu He, Xu~Tan, Yingce Xia, Di~He, Tao Qin, Zhibo Chen, and Tie-Yan Liu.
  2018.
\newblock Layer-wise coordination between encoder and decoder for neural
  machine translation.
\newblock In \emph{Advances in Neural Information Processing Systems}, pages
  7944--7954.

\bibitem[{Huang and Riloff(2011)}]{huang-riloff-2011-peeling}
Ruihong Huang and Ellen Riloff. 2011.
\newblock \href {https://www.aclweb.org/anthology/P11-1114} {Peeling back the
  layers: Detecting event role fillers in secondary contexts}.
\newblock In \emph{Proceedings of the 49th Annual Meeting of the Association
  for Computational Linguistics: Human Language Technologies}, pages
  1137--1147, Portland, Oregon, USA. Association for Computational Linguistics.

\bibitem[{Huang and Riloff(2012)}]{huang2012modeling}
Ruihong Huang and Ellen Riloff. 2012.
\newblock Modeling textual cohesion for event extraction.
\newblock In \emph{Twenty-Sixth AAAI Conference on Artificial Intelligence}.

\bibitem[{Ji and Grishman(2008)}]{ji-grishman-2008-refining}
Heng Ji and Ralph Grishman. 2008.
\newblock \href {https://www.aclweb.org/anthology/P08-1030} {Refining event
  extraction through cross-document inference}.
\newblock In \emph{Proceedings of ACL-08: HLT}, pages 254--262, Columbus, Ohio.
  Association for Computational Linguistics.

\bibitem[{Jia et~al.(2019)Jia, Wong, and Poon}]{jia-etal-2019-document}
Robin Jia, Cliff Wong, and Hoifung Poon. 2019.
\newblock \href {https://doi.org/10.18653/v1/N19-1370} {Document-level n-ary
  relation extraction with multiscale representation learning}.
\newblock In \emph{Proceedings of the 2019 Conference of the North {A}merican
  Chapter of the Association for Computational Linguistics: Human Language
  Technologies}, pages 3693--3704, Minneapolis, Minnesota. Association for
  Computational Linguistics.

\bibitem[{Jurafsky and Martin(2014)}]{jurafskyspeech}
Dan Jurafsky and James~H. Martin. 2014.
\newblock \href {https://www.worldcat.org/oclc/315913020} {\emph{Speech and
  language processing}}.
\newblock Prentice Hall, Pearson Education International.

\bibitem[{Kuhn(1955)}]{kuhn1955hungarian}
Harold~W Kuhn. 1955.
\newblock The hungarian method for the assignment problem.
\newblock \emph{Naval research logistics quarterly}, 2(1-2):83--97.

\bibitem[{Lample et~al.(2016)Lample, Ballesteros, Subramanian, Kawakami, and
  Dyer}]{lample-etal-2016-neural}
Guillaume Lample, Miguel Ballesteros, Sandeep Subramanian, Kazuya Kawakami, and
  Chris Dyer. 2016.
\newblock \href {https://doi.org/10.18653/v1/N16-1030} {Neural architectures
  for named entity recognition}.
\newblock In \emph{Proceedings of the 2016 Conference of the North {A}merican
  Chapter of the Association for Computational Linguistics: Human Language
  Technologies}, pages 260--270, San Diego, California. Association for
  Computational Linguistics.

\bibitem[{Li et~al.(2013)Li, Ji, and Huang}]{li-etal-2013-joint}
Qi~Li, Heng Ji, and Liang Huang. 2013.
\newblock \href {https://www.aclweb.org/anthology/P13-1008} {Joint event
  extraction via structured prediction with global features}.
\newblock In \emph{Proceedings of the 51st Annual Meeting of the Association
  for Computational Linguistics}, pages 73--82, Sofia, Bulgaria. Association
  for Computational Linguistics.

\bibitem[{Li et~al.(2015)Li, Nguyen, Cao, and
  Grishman}]{li-etal-2015-improving-event}
Xiang Li, Thien~Huu Nguyen, Kai Cao, and Ralph Grishman. 2015.
\newblock \href {https://doi.org/10.18653/v1/W15-4502} {Improving event
  detection with abstract meaning representation}.
\newblock In \emph{Proceedings of the First Workshop on Computing News
  Storylines}, pages 11--15, Beijing, China. Association for Computational
  Linguistics.

\bibitem[{Liao and Grishman(2010)}]{liao-grishman-2010-using}
Shasha Liao and Ralph Grishman. 2010.
\newblock \href {https://www.aclweb.org/anthology/P10-1081} {Using document
  level cross-event inference to improve event extraction}.
\newblock In \emph{Proceedings of the 48th Annual Meeting of the Association
  for Computational Linguistics}, pages 789--797, Uppsala, Sweden. Association
  for Computational Linguistics.

\bibitem[{Linguistic Data~Consortium(2005)}]{ace-2005}
(LDC) Linguistic Data~Consortium. 2005.
\newblock \href
  {https://www.ldc.upenn.edu/sites/www.ldc.upenn.edu/files/english-events-guidelines-v5.4.3.pdf}
  {English annotation guidelines for events}.
\newblock \emph{\texttt{https://www.ldc.upenn.edu/sites/\\
  www.ldc.upenn.edu/files/\\ english-events-guidelines-v5.4.3.pdf}}.

\bibitem[{Liu et~al.(2017)Liu, Chen, Liu, and Zhao}]{liu-etal-2017-exploiting}
Shulin Liu, Yubo Chen, Kang Liu, and Jun Zhao. 2017.
\newblock \href {https://doi.org/10.18653/v1/P17-1164} {Exploiting argument
  information to improve event detection via supervised attention mechanisms}.
\newblock In \emph{Proceedings of the 55th Annual Meeting of the Association
  for Computational Linguistics}, pages 1789--1798, Vancouver, Canada.
  Association for Computational Linguistics.

\bibitem[{Liu et~al.(2019)Liu, Huang, and Zhang}]{liu-etal-2019-open}
Xiao Liu, Heyan Huang, and Yue Zhang. 2019.
\newblock \href {https://doi.org/10.18653/v1/P19-1276} {Open domain event
  extraction using neural latent variable models}.
\newblock In \emph{Proceedings of the 57th Annual Meeting of the Association
  for Computational Linguistics}, pages 2860--2871, Florence, Italy.
  Association for Computational Linguistics.

\bibitem[{Liu et~al.(2018)Liu, Luo, and Huang}]{liu-etal-2018-jointly}
Xiao Liu, Zhunchen Luo, and Heyan Huang. 2018.
\newblock \href {https://doi.org/10.18653/v1/D18-1156} {Jointly multiple events
  extraction via attention-based graph information aggregation}.
\newblock In \emph{Proceedings of the 2018 Conference on Empirical Methods in
  Natural Language Processing}, pages 1247--1256, Brussels, Belgium.
  Association for Computational Linguistics.

\bibitem[{Luan et~al.(2018)Luan, He, Ostendorf, and
  Hajishirzi}]{luan-etal-2018-multi}
Yi~Luan, Luheng He, Mari Ostendorf, and Hannaneh Hajishirzi. 2018.
\newblock \href {https://doi.org/10.18653/v1/D18-1360} {Multi-task
  identification of entities, relations, and coreference for scientific
  knowledge graph construction}.
\newblock In \emph{Proceedings of the 2018 Conference on Empirical Methods in
  Natural Language Processing}, pages 3219--3232, Brussels, Belgium.
  Association for Computational Linguistics.

\bibitem[{Luan et~al.(2019)Luan, Wadden, He, Shah, Ostendorf, and
  Hajishirzi}]{luan-etal-2019-general}
Yi~Luan, Dave Wadden, Luheng He, Amy Shah, Mari Ostendorf, and Hannaneh
  Hajishirzi. 2019.
\newblock \href {https://doi.org/10.18653/v1/N19-1308} {A general framework for
  information extraction using dynamic span graphs}.
\newblock In \emph{Proceedings of the 2019 Conference of the North {A}merican
  Chapter of the Association for Computational Linguistics: Human Language
  Technologies}, pages 3036--3046, Minneapolis, Minnesota. Association for
  Computational Linguistics.

\bibitem[{Luo(2005)}]{luo-2005-coreference}
Xiaoqiang Luo. 2005.
\newblock \href {https://www.aclweb.org/anthology/H05-1004} {On coreference
  resolution performance metrics}.
\newblock In \emph{Proceedings of Human Language Technology Conference and
  Conference on Empirical Methods in Natural Language Processing}, pages
  25--32, Vancouver, British Columbia, Canada. Association for Computational
  Linguistics.

\bibitem[{Miwa and Bansal(2016)}]{miwa-bansal-2016-end}
Makoto Miwa and Mohit Bansal. 2016.
\newblock \href {https://doi.org/10.18653/v1/P16-1105} {End-to-end relation
  extraction using {LSTM}s on sequences and tree structures}.
\newblock In \emph{Proceedings of the 54th Annual Meeting of the Association
  for Computational Linguistics}, pages 1105--1116, Berlin, Germany.
  Association for Computational Linguistics.

\bibitem[{MUC-4(1992)}]{muc-1992-message}
MUC-4. 1992.
\newblock \href {https://www.aclweb.org/anthology/M92-1000} {Fourth message
  understanding conference ({MUC}-4)}.
\newblock In \emph{Proceedings of FOURTH MESSAGE UNDERSTANDING CONFERENCE
  ({MUC}-4)}, McLean, Virginia.

\bibitem[{Munkres(1957)}]{munkres1957algorithms}
James Munkres. 1957.
\newblock Algorithms for the assignment and transportation problems.
\newblock \emph{Journal of the society for industrial and applied mathematics},
  5(1):32--38.

\bibitem[{Nguyen et~al.(2016)Nguyen, Cho, and
  Grishman}]{nguyen-etal-2016-joint}
Thien~Huu Nguyen, Kyunghyun Cho, and Ralph Grishman. 2016.
\newblock \href {https://doi.org/10.18653/v1/N16-1034} {Joint event extraction
  via recurrent neural networks}.
\newblock In \emph{Proceedings of the 2016 Conference of the North {A}merican
  Chapter of the Association for Computational Linguistics: Human Language
  Technologies}, pages 300--309, San Diego, California. Association for
  Computational Linguistics.

\bibitem[{Nguyen and Grishman(2015)}]{nguyen-grishman-2015-event}
Thien~Huu Nguyen and Ralph Grishman. 2015.
\newblock \href {https://doi.org/10.3115/v1/P15-2060} {Event detection and
  domain adaptation with convolutional neural networks}.
\newblock In \emph{Proceedings of the 53rd Annual Meeting of the Association
  for Computational Linguistics and the 7th International Joint Conference on
  Natural Language Processing}, pages 365--371, Beijing, China. Association for
  Computational Linguistics.

\bibitem[{Patwardhan and Riloff(2009)}]{patwardhan-riloff-2009-unified}
Siddharth Patwardhan and Ellen Riloff. 2009.
\newblock \href {https://www.aclweb.org/anthology/D09-1016} {A unified model of
  phrasal and sentential evidence for information extraction}.
\newblock In \emph{Proceedings of the 2009 Conference on Empirical Methods in
  Natural Language Processing}, pages 151--160, Singapore. Association for
  Computational Linguistics.

\bibitem[{Peng et~al.(2017)Peng, Poon, Quirk, Toutanova, and
  Yih}]{peng-etal-2017-cross}
Nanyun Peng, Hoifung Poon, Chris Quirk, Kristina Toutanova, and Wen-tau Yih.
  2017.
\newblock \href {https://doi.org/10.1162/tacl_a_00049} {Cross-sentence n-ary
  relation extraction with graph {LSTM}s}.
\newblock \emph{Transactions of the Association for Computational Linguistics},
  5:101--115.

\bibitem[{Sundheim(1991)}]{sundheim-1991-overview}
Beth~M. Sundheim. 1991.
\newblock \href {https://www.aclweb.org/anthology/M91-1001} {Overview of the
  third {M}essage {U}nderstanding {E}valuation and {C}onference}.
\newblock In \emph{{T}hird {M}essage {U}understanding {C}onference ({MUC}-3):
  Proceedings of a Conference Held in {S}an {D}iego, {C}alifornia, {M}ay 21-23,
  1991}.

\bibitem[{Sundheim(1993)}]{sundheim-1993-message}
Beth~M. Sundheim. 1993.
\newblock \href {https://doi.org/10.3115/1119149.1119153} {The {M}essage
  {U}nderstanding {C}onferences}.
\newblock In \emph{TIPSTER TEXT PROGRAM: PHASE {I}: Proceedings of a Workshop
  held at Fredricksburg, Virginia, September 19-23, 1993}, pages 5--5,
  Fredericksburg, Virginia, USA. Association for Computational Linguistics.

\bibitem[{Sutskever et~al.(2014)Sutskever, Vinyals, and
  Le}]{sutskever2014sequence}
Ilya Sutskever, Oriol Vinyals, and Quoc~V Le. 2014.
\newblock Sequence to sequence learning with neural networks.
\newblock In \emph{Advances in neural information processing systems}, pages
  3104--3112.

\bibitem[{Wadden et~al.(2019)Wadden, Wennberg, Luan, and
  Hajishirzi}]{wadden-etal-2019-entity}
David Wadden, Ulme Wennberg, Yi~Luan, and Hannaneh Hajishirzi. 2019.
\newblock \href {https://doi.org/10.18653/v1/D19-1585} {Entity, relation, and
  event extraction with contextualized span representations}.
\newblock In \emph{Proceedings of the 2019 Conference on Empirical Methods in
  Natural Language Processing (EMNLP-IJCNLP)}, pages 5784--5789, Hong Kong,
  China. Association for Computational Linguistics.

\bibitem[{Walker et~al.(2006)Walker, Strassel, Medero, and
  Maeda}]{walker2006ace}
Christopher Walker, Stephanie Strassel, Julie Medero, and Kazuaki Maeda. 2006.
\newblock Ace 2005 multilingual training corpus.
\newblock \emph{Linguistic Data Consortium, Philadelphia}, 57.

\bibitem[{Wang and Poon(2018)}]{wang-poon-2018-deep}
Hai Wang and Hoifung Poon. 2018.
\newblock \href {https://doi.org/10.18653/v1/D18-1215} {Deep probabilistic
  logic: A unifying framework for indirect supervision}.
\newblock In \emph{Proceedings of the 2018 Conference on Empirical Methods in
  Natural Language Processing}, pages 1891--1902, Brussels, Belgium.
  Association for Computational Linguistics.

\bibitem[{Yang and Mitchell(2016)}]{yang-mitchell-2016-joint}
Bishan Yang and Tom~M. Mitchell. 2016.
\newblock \href {https://doi.org/10.18653/v1/N16-1033} {Joint extraction of
  events and entities within a document context}.
\newblock In \emph{Proceedings of the 2016 Conference of the North {A}merican
  Chapter of the Association for Computational Linguistics: Human Language
  Technologies}, pages 289--299, San Diego, California. Association for
  Computational Linguistics.

\bibitem[{Yao et~al.(2019)Yao, Ye, Li, Han, Lin, Liu, Liu, Huang, Zhou, and
  Sun}]{yao2019docred}
Yuan Yao, Deming Ye, Peng Li, Xu~Han, Yankai Lin, Zhenghao Liu, Zhiyuan Liu,
  Lixin Huang, Jie Zhou, and Maosong Sun. 2019.
\newblock \href {https://doi.org/10.18653/v1/P19-1074} {{D}oc{RED}: A
  large-scale document-level relation extraction dataset}.
\newblock In \emph{Proceedings of the 57th Annual Meeting of the Association
  for Computational Linguistics}, pages 764--777, Florence, Italy. Association
  for Computational Linguistics.

\bibitem[{Zhang et~al.(2019)Zhang, Ji, and Sil}]{zhangtongtao2019joint}
Tongtao Zhang, Heng Ji, and Avirup Sil. 2019.
\newblock \href
  {http://www.data-intelligence-journal.org/static/publish/95/08/A8/2FAD3A453BA99B7D61557655B7/Tongtao_Zhang.pdf}
  {Joint entity and event extraction with generative adversarial imitation
  learning}.
\newblock \emph{Data Intelligence}, 1(2):99--120.

\bibitem[{Zhao et~al.(2018)Zhao, Jin, Wang, and
  Cheng}]{zhao-etal-2018-document}
Yue Zhao, Xiaolong Jin, Yuanzhuo Wang, and Xueqi Cheng. 2018.
\newblock \href {https://doi.org/10.18653/v1/P18-2066} {Document embedding
  enhanced event detection with hierarchical and supervised attention}.
\newblock In \emph{Proceedings of the 56th Annual Meeting of the Association
  for Computational Linguistics}, pages 414--419, Melbourne, Australia.
  Association for Computational Linguistics.

\end{thebibliography}
\bibliographystyle{acl_natbib}

\cleardoublepage
\newpage
\appendix

\section{Appendices}
\label{sec:appendix}


\subsection{CEAF-REE metric}
\paragraph{Notations}
First we provide the necessary notations. 
Let reference (gold) role-filler entities of one role in a document $d$ be:
\begin{equation}
\nonumber
R(d) = \{R_i: i=1,2,..., |R(d)|\}
\end{equation}

and predicted role-filler entities be:
\begin{equation}
\nonumber
S(d) = \{S_i: i=1,2,..., |S(d)|\}
\end{equation}

Let $m$ be the smaller one of $|R(d)|$ and $|S(d)|$, i.e., $m = min (|R(d)|, |S(d)|)$.
Let $R_m \subset R$ and $S_m \subset S$ be any subsets with $m$ entities.
Let $G(R_m, S_m)$ be the set of \textit{one-to-one} entity maps from $R_m$ to $S_m$,
and $G_m$ be the set of all possible one-to-one maps (of size-$m$) between subsets of $R$ and $S$.
Obviously, we have $G(R_m, S_m) \in G_m$.

The similarity function $\phi (r,s)$ measures the ``similarity'' between two entities. It takes non-negative values: zero-value means role-filler entity $r$ is not subset of $s$.
\begin{equation}
\nonumber
\phi(r, s)=
\left\{
\begin{aligned}
1 &, \ \ \text{if} \ \ s \subseteq r \\
0 &, \ \ \text{otherwise} \\
\end{aligned}
\right.
\end{equation}

\paragraph{Calculating CEAF-REE score}

Next we present how to calculate the CEAF-REE score. Given the document $d$, for a certain event role (e.g., \target), with its gold entities $R$ and system predicted entities $S$, we first find the best alignment $g^*$ by maximizing the total similarity $\Phi$ (maximum bipartite matching algorithm is applied in this step):
\begin{equation}
\nonumber
g^* = \text{arg} \max_{g \in G_m} \Phi (g) \\
= \text{arg} \max_{g \in G_m} \sum_{r \in R_{m}} \phi (r, g(r))
\end{equation}

Let $R^*_m$ and $S^*_m = g^*(R^*_m)$ denote the gold and predicted role-filler entity subset (respectively), where best matching $g^*$ is obtained. Then the maximum total similarity is,
\begin{equation}
\nonumber
\Phi (g^*) = \sum_{r \in R_{m}^*} \phi (r, g^*(r))
\end{equation}

\begin{figure}[!ht]
\centering
\resizebox{\columnwidth}{!}{\includegraphics[]{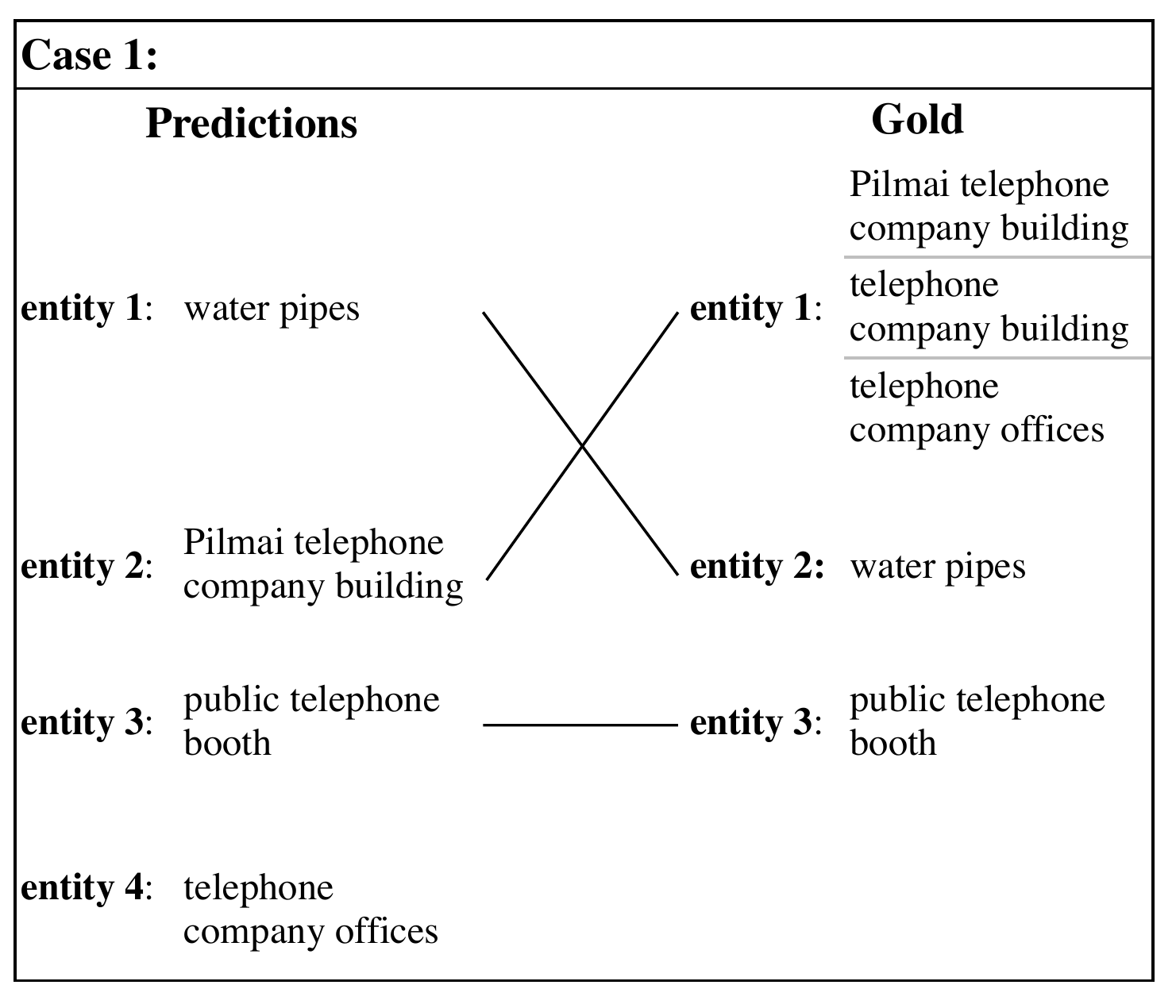}}
\resizebox{\columnwidth}{!}{\includegraphics[]{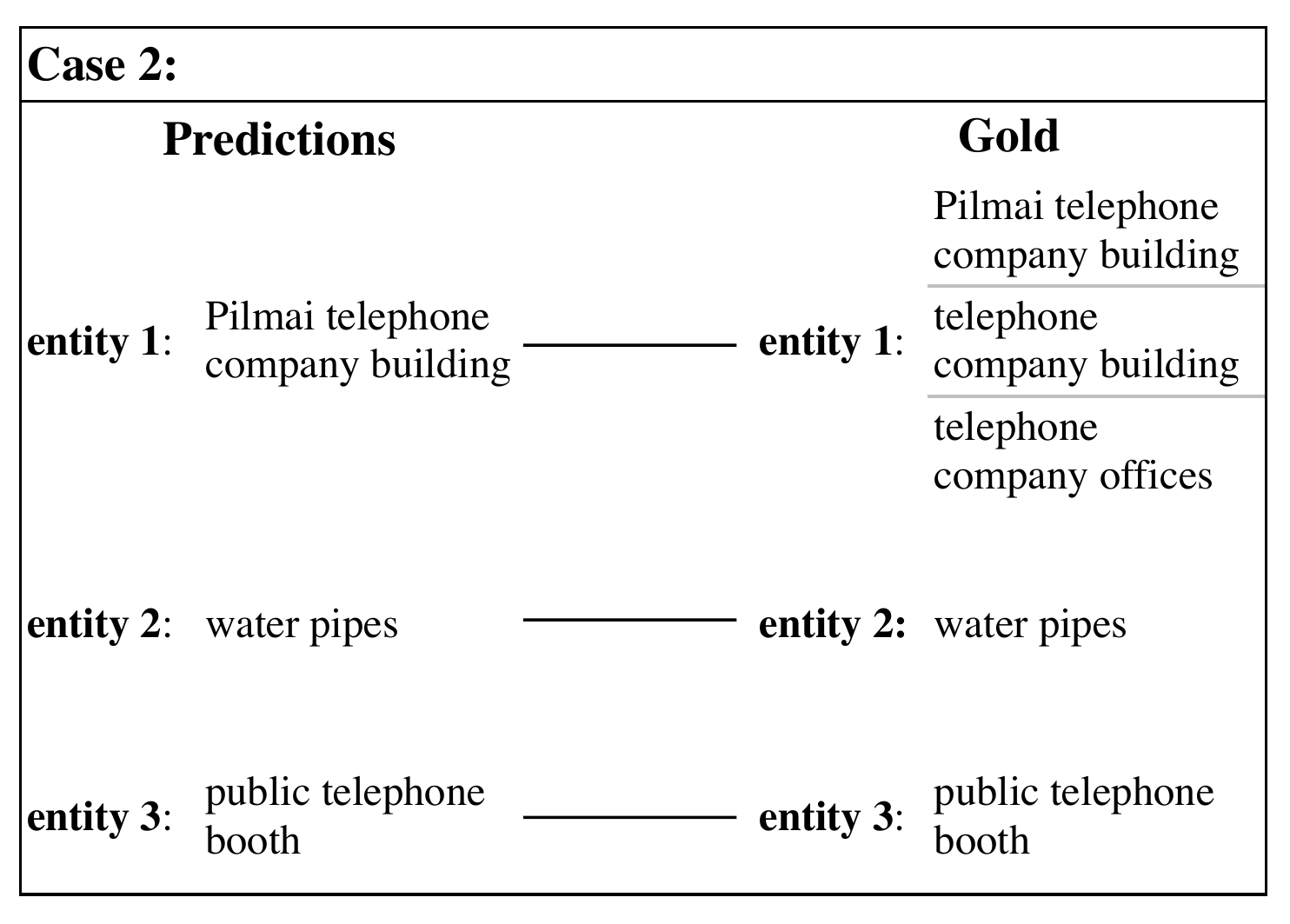}}
\resizebox{\columnwidth}{!}{\includegraphics[]{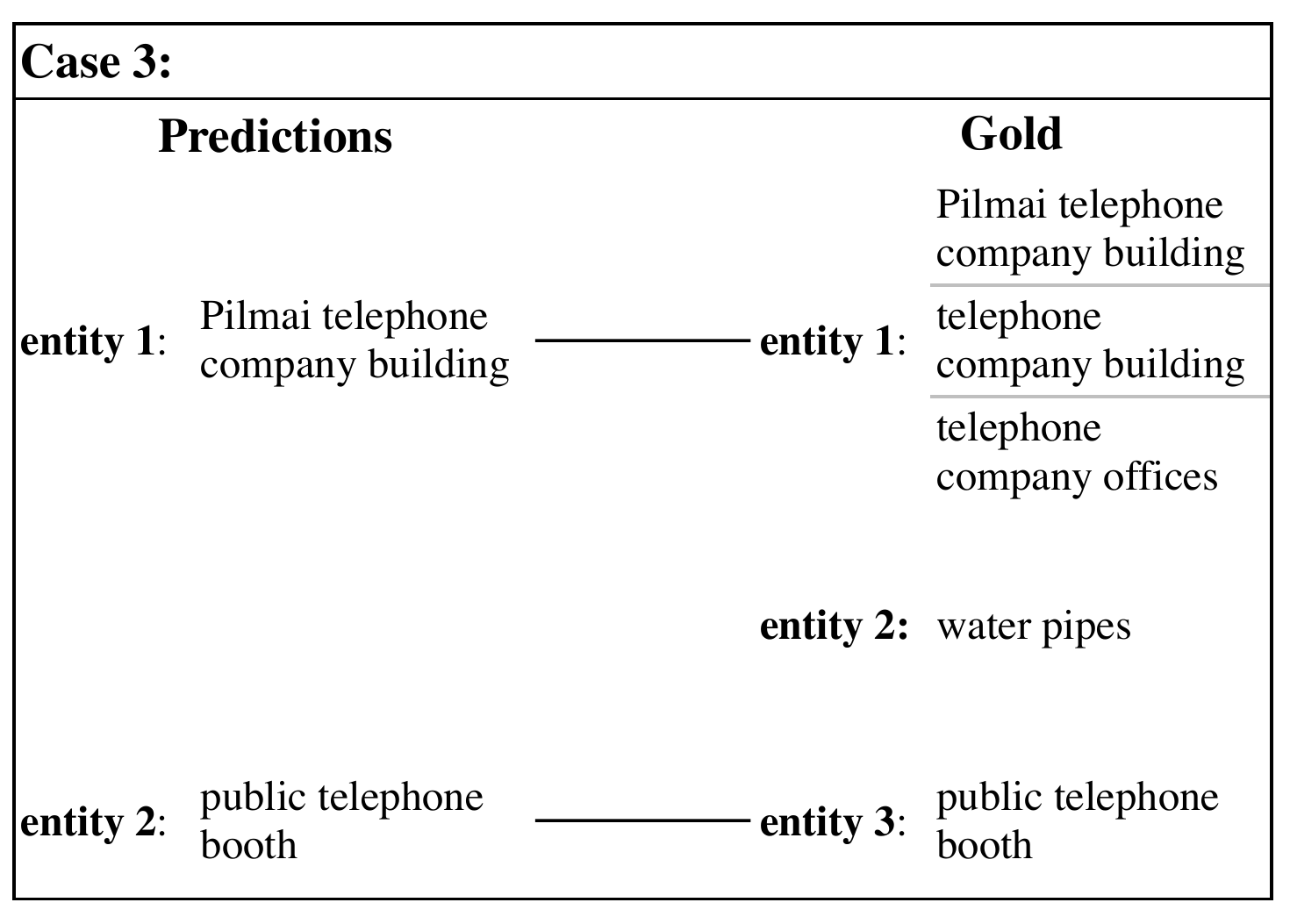}}
\caption{Test cases for the CEAF-REE metric (The lines between entities denote the best alignment/matching).}
\label{fig:test_case}
\end{figure}

\begin{table}[!ht]
\centering
\small
\begin{tabular}{l|ccc}
\toprule
       & Precision & Recall & F1   \\\midrule
case 1 & 0.75      & 1.00   & 0.86 \\
case 2 & 1.00      & 1.00   & 1.00 \\
case 3 & 1.00      & 0.67   & 0.80 \\
\bottomrule
\end{tabular}
\caption{CEAF-REE scores for case-$\{1,2,3\}$.}
\label{tab:test_scores}
\end{table}

We can also calculate the entity self-similarity with $\phi$.
Finally, we calculate the precision, recall and F-measure for CEAF-REE as follows:
\begin{equation}
\nonumber
\begin{gathered}
    prec = \frac{\Phi (g^*)}{\sum_{i}\phi(S_i, S_i)}\\
    recall = \frac{\Phi (g^*)}{\sum_{i}\phi(R_i, R_i)}\\
    F = \frac{2*prec*recall}{prec+recall}
\end{gathered}
\end{equation}

We list several cases (Figure~\ref{fig:test_case}) and their CEAF-REE scores (Table~\ref{tab:test_scores}) to facilitate understanding. 

For more details, readers can refer to Section 2 of \newcite{luo-2005-coreference}.



\subsection{Others}

\paragraph{Code and Computing}
We use the NVIDIA TITAN Xp GPU for our computing infrastructure.
We build our model based on the Hugging-face NER models' implementation \url{https://github.com/huggingface/transformers/tree/3ee431dd4c720e67e35a449b453d3dc2b15ccfff/examples/ner}. 
The hyperparameters can also be obtained from the default values in the repo.

\paragraph{Link to Corpus}
The raw corpus and preprocessing script can be found at: \url{https://github.com/brendano/muc4_proc}

\paragraph{Dependencies}
\begin{itemize}
    \item Python 3.6.10
    \item Transformers: transformers 2.4.1 installed from source.
    \item Pytorch-Struct: Install from Github.
    \item Pytorch-Lightning==0.7.1
    \item Pytorch==1.4.0
\end{itemize}

\end{document}